\documentclass[conference]{IEEEtran}
\IEEEoverridecommandlockouts

\usepackage{times}
\usepackage{hyperref}
\usepackage{url}

\usepackage{pbox}
\usepackage{enumitem}
\usepackage{array}
\usepackage{makecell}
\usepackage{booktabs}
\usepackage{subcaption}
\usepackage{wrapfig}

\usepackage{amsmath,amssymb,amsfonts}%

\usepackage{tabularx}
\usepackage{fancyvrb}
\usepackage[dvipsnames]{xcolor}

\usepackage{arydshln} % for \hdashline in table 3
\usepackage{booktabs} % for \midrule
\usepackage[capitalise]{cleveref} % for cref

\usepackage{cite}
\usepackage{algorithmic}

\usepackage{textcomp}

\newcommand{\cf}{cf.\ }
\newcommand{\ie}{i.\,e.\ }

\newcommand{\wrt}{w.\,r.\,t.\ }

\newcommand{\RSLM}{RSLM\xspace}

\makeatletter
\def\adl@drawiv#1#2#3{%
        \hskip.5\tabcolsep
        \xleaders#3{#2.5\@tempdimb #1{1}#2.5\@tempdimb}%
                #2\z@ plus1fil minus1fil\relax
        \hskip.5\tabcolsep}
\newcommand{\cdashlinelr}[1]{%
  \noalign{\vskip\aboverulesep
           \global\let\@dashdrawstore\adl@draw
           \global\let\adl@draw\adl@drawiv}
  \cdashline{#1}
  \noalign{\global\let\adl@draw\@dashdrawstore
           \vskip\belowrulesep}}
\makeatother
\def\BibTeX{{\rm B\kern-.05em{\sc i\kern-.025em b}\kern-.08em
    T\kern-.1667em\lower.7ex\hbox{E}\kern-.125emX}}

\usepackage{amsmath,amsfonts,bm}

\def\eqref#1{equation~\ref{#1}}
\def\1{\bm{1}}

\DeclareMathAlphabet{\mathsfit}{\encodingdefault}{\sfdefault}{m}{sl}
\SetMathAlphabet{\mathsfit}{bold}{\encodingdefault}{\sfdefault}{bx}{n}

\usepackage{graphicx}
\usepackage{xspace}
\usepackage{multirow}

\title{Radar Spectra-Language Model for Automotive Scene Parsing}

\begin{document}

\author{
    \IEEEauthorblockN{
        Mariia Pushkareva\textsuperscript{1},
        Yuri Feldman\textsuperscript{1}\textsuperscript{\textdagger},
        Csaba Domokos\textsuperscript{2},
        Kilian Rambach\textsuperscript{2},
        Dotan Di Castro\textsuperscript{1}
    }\\
    \IEEEauthorblockA{
        \textit{Bosch Center for Artificial Intelligence},
        Haifa, Israel\textsuperscript{1} and Renningen, Germany\textsuperscript{2}\\
        \textsuperscript{\textdagger}Corresponding author. Email: yuri.feldman@il.bosch.com
    }
}

\maketitle

% !TeX root = root.tex
\begin{abstract}
Radar sensors are low cost, long-range, and weather-resilient.
Therefore, they are widely used for driver assistance functions, and are expected to be crucial for the success of autonomous driving in the future.
In many perception tasks only pre-processed radar point clouds are considered.
In contrast, radar spectra are a raw form of radar measurements and contain more information than radar point clouds.
However, radar spectra are rather difficult to interpret.
In this work, we aim to explore the semantic information contained in spectra in the context of automated driving, thereby moving towards better interpretability of radar spectra.
To this end, we create a radar spectra-language model, allowing us to query radar spectra measurements for the presence of scene elements using free text.
We overcome the scarcity of radar spectra data by matching the embedding space of an existing vision-language model.
Finally, we explore the benefit of the learned representation for scene retrieval using radar spectra only, and obtain improvements in free space segmentation and object detection merely by injecting the spectra embedding into a baseline model.
\end{abstract}


\section{Introduction}
\label{intro}
\begin{figure}
    \centering
        \includegraphics[page=1, width=\linewidth, trim={2.3cm 4.61cm 14.45cm 3.3cm},  clip]{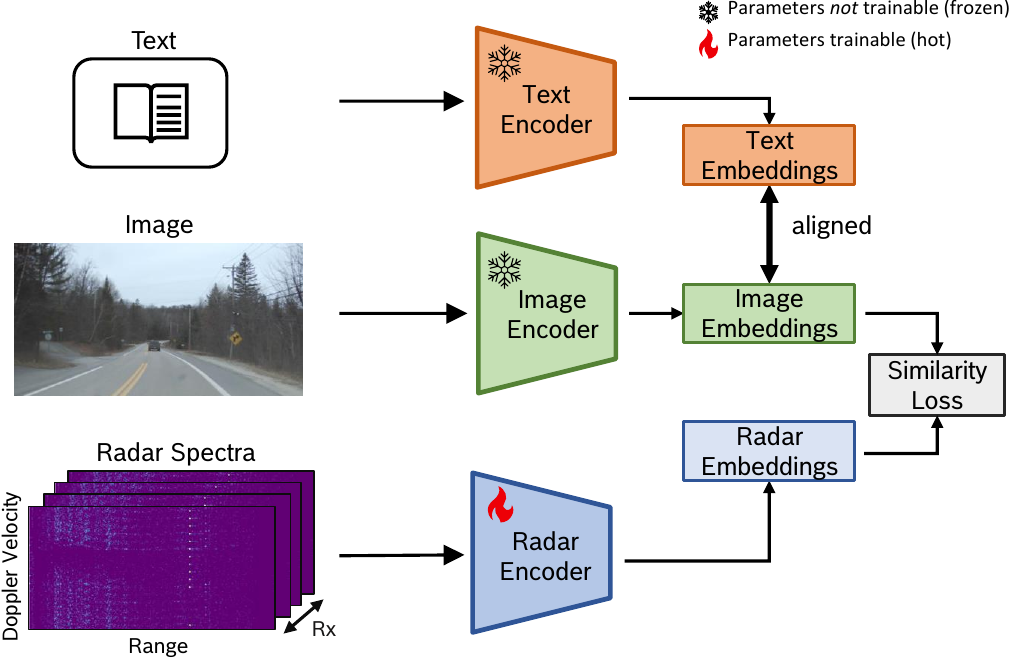}
        %trim: left lower right upper
\centering
    \caption{Training of a radar spectra-language model utilizes a frozen vision-language model for supervision. Radar spectra encoder is trained to match image embeddings of the corresponding RGB images. In this way, text embeddings get aligned to radar embeddings as well.}
    \label{fig:train_radar_encoder}
    \vspace{-0.5cm}
\end{figure}

Radar is a valuable sensing modality in the automotive domain, combining the benefits of low hardware cost with long-range and weather-resilient sensing.
Radar sensors are already used for driver assistance functions and are expected to be crucial for autonomous driving.
Nevertheless, developing a well performing perception algorithm, e.g., to detect all relevant objects, is a challenging task.
Numerous radar perception algorithms are based on radar point cloud data as input \cite{Ulrich2022pillars,Danzer2019pointnet,Svenningsson2021radarGNN,bang2024radardistill}. To compute the point cloud data, first the measured baseband time signal is converted to radar spectra.
Then local intensity maxima, the radar reflections, are filtered out. This results in a list of radar reflections, the radar point cloud. Therefore, information that is available in the raw spectral radar data, is inevitably lost in the point cloud data \cite{Yao23arxiv-DataRepresentations}.
Recent work \cite{Major_2019_ICCV,Cozma2021DeepHybrid,Wang2021rodnet,paek-2022-k-radar-dataset,Rebut2021radial,Zhang2021raddet} shows that perception algorithms applied on radar spectra can achieve improved performance.
Nevertheless, working on radar spectra introduces new challenges:
First of all, there are only a small number of labeled datasets available providing radar spectra.
Furthermore, radar spectra data is difficult to interpret by humans, as evidenced by \cref{fig:train_radar_encoder} depicting the RGB image alongside its corresponding range-Doppler spectrum.
This leads naturally to the question: what scene information is captured in radar spectra?

To address the above, we propose an approach which a) does not need any labeled radar spectra data and b) can be used to query radar spectra for contents of interest using free text.
This is a step towards understanding the semantic content of radar spectra. 
We propose to train a radar spectra-language model (\RSLM) for automotive scenarios, motivated by the tremendous success of vision-language models (VLMs) like CLIP \cite{Radford2021clip}, or DALL-E \cite{Ramesh2021dalle}.

To train the radar encoder, we utilize the frozen image encoder of a VLM, \ie the weights of the image encoder are not adapted during training of the radar encoder, \cf \cref{fig:train_radar_encoder}.
During training, the radar encoder embeddings are forced to match the feature embeddings of the image encoder.
In the VLM, the output feature embeddings of the text encoder are aligned to the feature embeddings of the image encoder, \ie text domain is connected to image domain.
By aligning the feature embeddings of the radar encoder to the ones of the image encoder, the feature embeddings of the radar encoder are aligned to the ones of the text encoder as well, \ie text domain is connected to radar spectra domain, see \cref{fig:train_radar_encoder}.
In this way, we construct the \textit{radar spectra-language model}.
To the best of our knowledge, we are the first ones to train a radar-language model.
Note that for training the radar encoder only paired image-radar spectra samples are needed, no labeled spectral radar data is necessary.
This tackles the problem of a large labeled radar spectra dataset, which is usually needed for a supervised training.

We are especially interested in automotive applications.
Since performance of off-the-shelf VLMs is not satisfactory, we fine-tune VLMs for automotive scenes.
To explore the semantic content of radar spectra, we benchmark the \RSLM on scene retrieval tasks: Free text is used to describe a scene, and the \RSLM is used to search for data samples which fit to this scene description.
Moreover, we show that the \RSLM can be used to improve the performance on two downstream tasks, object detection on radar spectra and free space space estimation.

Our main contributions can be summarized as follows:\\
1) We propose training and evaluation of the first radar spectra-language model.
2) We benchmark scene retrieval using the radar spectra-language model, exploring semantic content of radar spectra.
3) We investigate the benefits of the learned radar feature embeddings on two downstream tasks: object detection and free space estimation.

\section{Related Work}
\label{related_work}
\paragraph*{Vision-language Models}
Large VLMs have shown great potential in learning representations that are transferable across a wide range of downstream tasks.
An efficient way to learn image representations by making use of contrastive training on image-caption pairs was proposed in \cite{Radford2021clip}.
\cite{Zhai2022lit} shows the advantage of fine-tuning text models with frozen (pre-trained) image models.
However, the connection between text and other modalities has received less attention.
\cite{Girdhar23imagebind} proposed to train encoders of several modalities.
The closest approach to our work is LidarCLIP~\cite{hess2023lidarclip}, which learns a mapping from Lidar point clouds to CLIP~\cite{Radford2021clip} embedding space, effectively relating text and Lidar data through the image domain.
Our work was inspired by that idea, however, we consider a new input modality, namely \textit{radar spectra}.
We leverage vision-language models to achieve a better representation for radar spectra input.

\paragraph*{VLMs for Automotive Scene Understanding}
In recent research, VLMs are used for automotive applications \cite{Zhou23arxiv-VisionLM_survey, Yang23arxiv-LLM_AD_survey}.
Scene understanding with VLMs is investigated in the form of object detection \cite{Najibi23arxiv-3d_perception_distillation, Ding23arxiv-HiLMD} and visual questioning answering (VQA) \cite{Qian23arxiv-NuScenesQA, Dewangan23arxiv-Talk2BEV}, producing usually descriptions which capture only a subset of scene elements.
Captioning approaches \cite{Ding23arxiv-HiLMD} require expensive ground truth annotation and \cite{hess2023lidarclip} relies on a large-scale automotive dataset. Both are not available in our case.
Romero et al.~\cite{Romero23arxiv-Zelda} propose an approach, that matches the input scene measurement to an embedding vector, which lies in the same representation space as the text embedding. Thus the model can be queried using free text.
However, it utilizes out-of-the-box CLIP, and is limited by its performance.

\paragraph*{Object Detection on Spectra}
Object detection on automotive radar spectra is attracting increasing interest since recent introduction of public datasets \cite{Rebut2021radial,Wang2021rodnet,paek-2022-k-radar-dataset}.
A radar dataset and a one-stage detector generating both 3D and 2D bounding boxes was proposed in \cite{Zhang2021raddet}.
The CRUW dataset and an object detection network on range-azimuth radar spectra was presented in \cite{Wang2021rodnet}.
FFT-RadNet \cite{Rebut2021radial} eliminates the overhead of computing the range-azimuth-Doppler tensor by learning to recover angles from a range-Doppler spectrum. 
DAROD \cite{Decourt2022darod} is an adaptation of Faster R-CNN for automotive radar on range-Doppler spectra.
\cite{Giroux2023TFFTRadNet} proposed hierarchical Swin vision transformers for radar object detection.

\section{Proposed Approach}
\label{method}
Since the introduction of VLMs \cite{Radford2021clip}, the coupling of image and text latent representation spaces has been leveraged to enable semantic perception ``in-the-wild" accross modalities \cite{Girdhar23imagebind}. 
We aim to harness this generalization ability to examine the semantic content of radar spectra of automotive scenes.
To this end, we train a radar spectra-language model, consisting of a spectra encoder and a text encoder with a shared embedding space, representing the observed scenes.
The radar spectra encoder is trained by using paired radar spectra-image measurements from automotive driving datasets. It is trained to match the embedding space of a VLM, following \cite{hess2023lidarclip}.
To obtain an embedding space that better fits our data we formulate a process for fine-tuning a VLM using generated captions of automotive scenes, without any human annotations.
We validate our approach in two ways:
1) We evaluate our method on a spectra retrieval task using text queries, directly attempting to shed light on what elements of the scene are captured in radar spectra.
2) We inject the spectra embedding in a baseline object-detection and segmentation network to observe an improvement in detection and segmentation performance.

The rest of this section is organized as follows: The proposed process of fine-tuning a VLM with automotive data is described in \cref{sec:clip_finetuning}. We explain the training and architecture of the proposed radar spectra encoder in \cref{sec:train_encoder}, and present its application for downstream tasks in \cref{sec:method_detection}.

\subsection{VLM Fine-tuning}
\label{sec:clip_finetuning}
Publicly available VLMs are generally not specifically adapted to automotive scenes, e.g., CLIP accuracy of zero-shot classification for KITTI dataset  varies from 21\% to 44\% \cite{Radford2021clip}.
Therefore, we fine-tune a baseline VLM for road scenes.
We use a segmentation model to generate labels for the presence and position of different objects within each image.
Using these labels, multiple different captions are generated for each dataset frame, based on the objects which are present in this frame. This way, diverse captions can be generated automatically. Those captions along with the corresponding images are used to fine-tune the VLM.

\subsection{Radar Spectra-language Model}
\label{sec:train_encoder}
\begin{figure}
    \centering
    \includegraphics[width=\linewidth, trim={2.7cm 4.8cm 4.2cm 0.5cm}, clip]{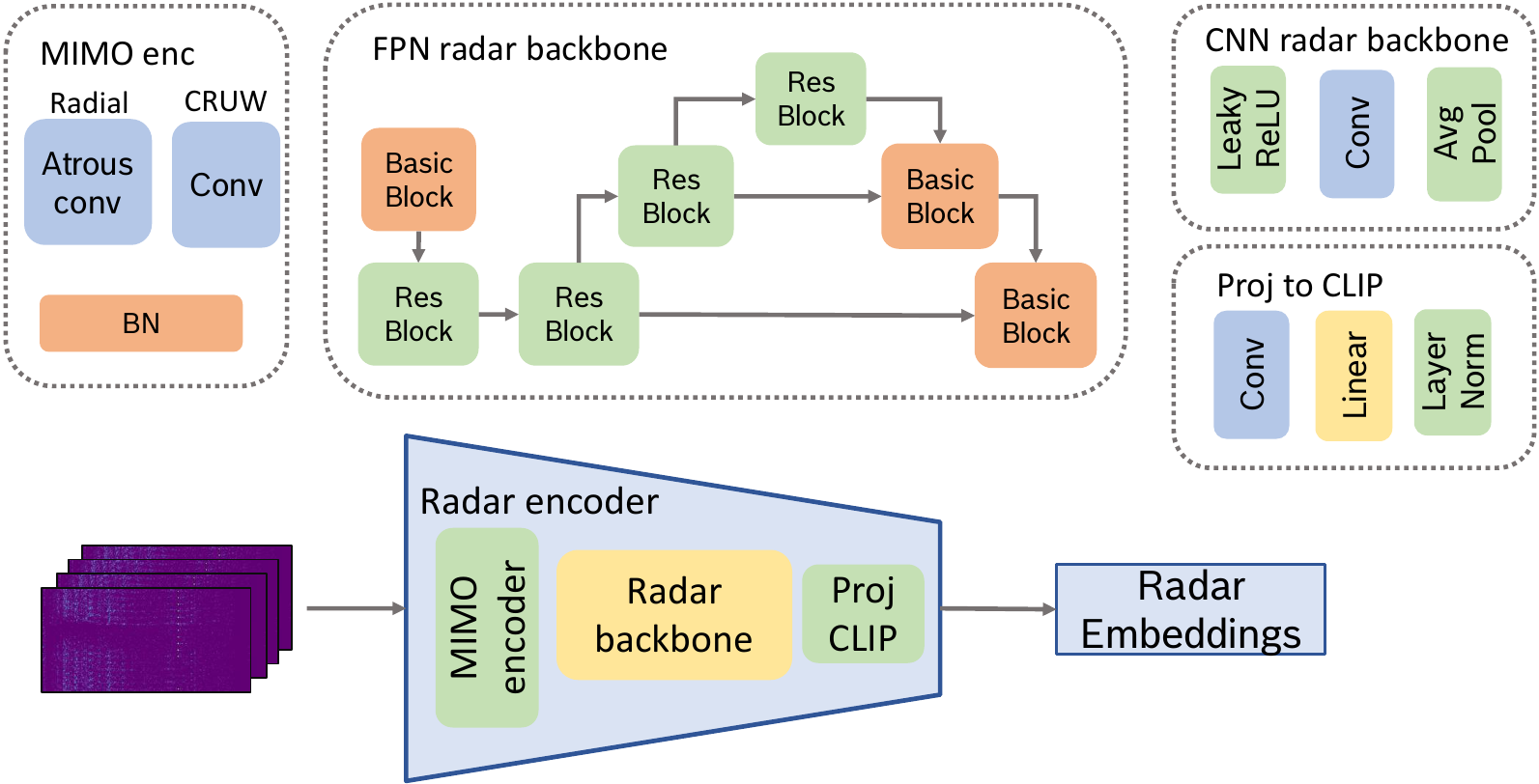}
        %trim: left lower right upper
    \caption{Architecture of the radar encoder, with FPN or CNN radar backbone. The MIMO encoder is chosen according to the dataset (CRUW or RADIal).}
    \label{fig:radarEncoder_architecture}
    \vspace{-0.5cm}
\end{figure}

To obtain paired radar-spectra and text encoders we use a similar concept as presented in \cite{hess2023lidarclip}.
We train a radar encoder to output similar embeddings to a VLM image encoder. During training of the radar encoder, the VLM model is frozen, \ie the weights of the VLM model are fixed, see \cref{fig:train_radar_encoder}.
Matching pairs of radar spectra and images are input to the network: an image to the frozen VLM image encoder and a corresponding radar spectrum to the radar encoder.
We train the radar encoder to minimize the difference between the image and radar encoder outputs, where both outputs, the radar embeddings and image embeddings, have the same size.
In this work, we compare two networks for the radar encoder: A network with a CNN backbone and a network with a Feature Pyramid Network (FPN) backbone, \cf \cref{fig:radarEncoder_architecture}.

\textbf{CNN radar encoder}
\label{sec:cnn_encoder} 
The CNN network includes three convolutional layers, batch-normalization, average pooling, a fully-connected layer and a layer normalization.
The parameters of the first convolutional layer depend on the radar spectrum type and the number of input channels of the dataset at hand. 

For the RADIal dataset, which includes range-Doppler spectra, we use the recommended paramters given by \cite{Rebut2021radial}.
For the CRUW dataset, which consists of range-azimuth spectra, the parameters are chosen according to the spectra dimensions.

\textbf{FPN radar encoder}
We choose the Feature Pyramid Network (FPN) of FFT-RadNet \cite{Rebut2021radial} as the radar encoder. Detection, and segmentation heads are not included in the radar encoder. 
A convolutional layer and fully-connected layer are added, to project the output to the same space as the CLIP embeddings.
The parameters of the first convolutional layer in the radar encoder are the same as for the CNN radar encoder, and depend on the dataset at hand.

\subsection{Downstream Tasks}
\label{sec:method_detection}
\begin{figure}
    \centering
    \includegraphics[page=1, width=\linewidth, clip, trim={3cm 5cm 7cm 5cm}]{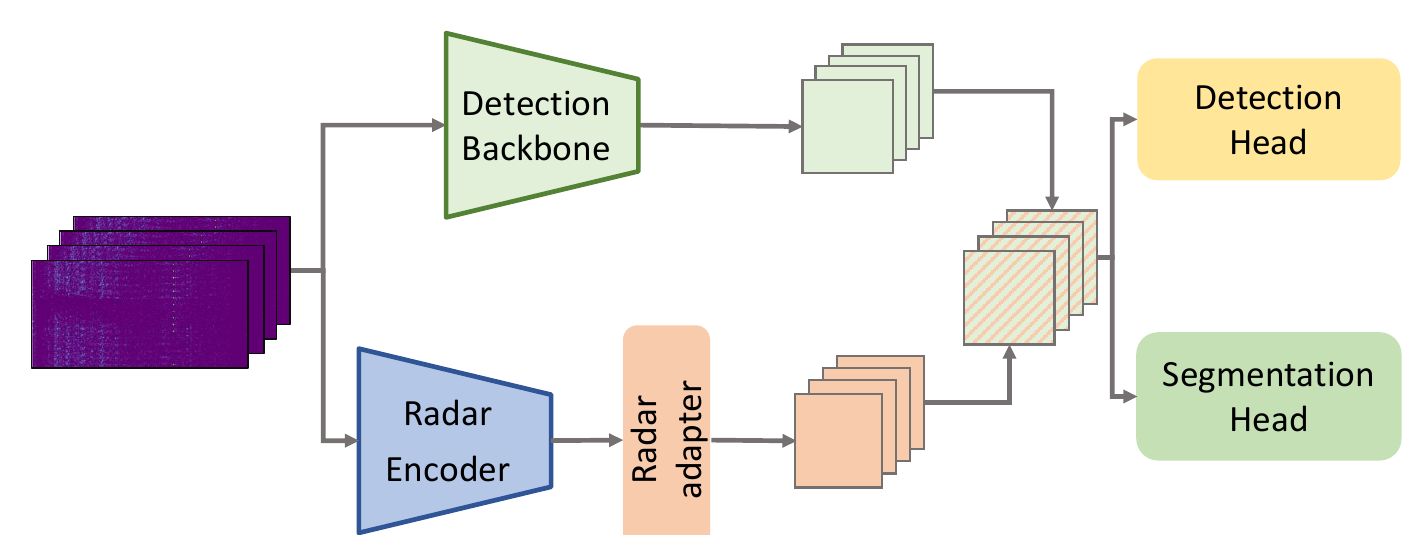}
        %trim: left lower right upper
    \centering
    \caption{\RSLM-Aided detection and segmentation architecture. Input spectra are concurrently fed into the detection backbone and the radar encoder from the pre-trained \RSLM.}
    \label{fig:rslm_detector}
    \vspace{-0.5cm}
\end{figure}

To investigate the benefit of the learned radar spectra feature embeddings for different downstream tasks, we consider object detection as well as free space estimation as two applications.
To this end, we combine the trained radar encoder with a detection and segmentation network. The overall architecture is depicted in \cref{fig:rslm_detector}. We propose to inject embeddings from the pre-trained \RSLM radar encoder into the detection network.
We hypothesize that this would introduce a semantic prior benefiting detection and segmentation. Below, we provide details on this architecture and the loss function for its training.

\textbf{Detection Backbone}
\label{sec:detector}
We choose FFT-RadNet as used by \cite{Rebut2021radial} as our detection backbone. 
There exists an optimized version of FFT-RadNet \cite{Yang2023ADCNet}, but hyperparameters haven't been made public. T-FFTRadNet by \cite{Giroux2023TFFTRadNet} uses a Swin \cite{Liu21iccv-Swin} backbone as opposed to FPN in FFT-RadNet. Since code and exact parameters are not available for T-FFTRadNet, we have choosen FFT-RadNet as baseline.
We use both the detection and the driveable space segmentation heads, as defined in \cite{Rebut2021radial}.

\textbf{Incorporating the \RSLM Embeddings} 
\label{sec:det_clip}
The input radar spectra tensor is concurrently fed into the detection backbone and the radar encoder.
The radar feature embeddings output by the radar encoder are transformed by an adapter branch to match the size of the output features of the detection backbone, and are summed with those.
As radar encoder we use the FPN radar encoder. 

\textbf{Loss}
The loss function is defined as $L = L_\text{det} + \lambda L_\text{seg}$,
where $0<\lambda\in\mathbb{R}$ is a weighting factor and the detection and segmentation losses are defined as follows:
\begin{gather}
    L_\text{det} = \text{focal}(y_\text{class}, \hat{y}_\text{class}) + \beta~\text{smooth-}L_1 (y_\text{reg} - \hat{y}_\text{reg}),
    \label{eq:detect_loss} \\
    L_\text{seg} = \sum_{r, a} \text{BCE}(y_\text{seg}(r, a), \hat{y}_\text{seg}(r, a)),
    \label{eq:segm_loss}
\end{gather}
where $0<\beta\in\mathbb{R}$; $\text{focal}(y_\text{class}, \hat{y}_\text{class})$ is the focal loss for true $y_\text{class}$ and predicted $\hat{y}_\text{class}$ class labels; smooth-$L_1$ is the smooth $L_1$ loss, where $y_\text{reg}$ and $\hat{y}_\text{reg}$ denote the ground-truth polar coordinates of the centers of the bounding boxes and the predicted ones, respectively.
$\text{BCE}$ denotes the binary cross entropy loss for true free-space map $y_\text{seg}$ and predicted map $\hat{y}_\text{seg}$, where $r$ and $a$ stand for range and azimuth coordinates, respectively.

\section{Experiments}
\label{experiments}
%%------------ CLIP finetuning -------------
We present experimental results for \textit{fine-tuning the CLIP image encoder} for automotive scenarios.
The semantic content of radar spectra is analyzed in \cref{sec:experiment_radar_encoder}, where the \RSLM is evaluated on a \textit{retrieval} task. In \cref{sec:experiment_object_detection} the benefit of the trained radar encoder is evaluated on two downstream tasks: \textit{radar object detection} and \textit{free space space estimation}.

\subsection{Datasets}
\label{sec:datasets}
In this paper we use three datasets for autonomous driving: 
1) RADIal \cite{Rebut2021radial} has 8252 annotated frames, each with a range-Doppler spectrum of size $512 \times 256 \times 16$, Lidar, radar reflections, images, centers of cars, and free space annotations in birds-eye view.
2) The CRUW dataset \cite{Wang2021rodnet} has 40,734 annotated frames with range-azimuth spectra of size $128 \times 128 \times 8$, RGB images, centers of \textit{cars}, \textit{pedestrians} and \textit{cyclists} in range-azimuth coordinates.
3) The nuScenes dataset \cite{Caesar2020nuscenes} includes Lidar and radar point clouds, camera, IMU and GPS data.
In this work, we only use images from the train-validation split with 40,157 samples.

\subsection{VLM finetuning}
\label{sec:experiment_clip_finetuning}
We use Open CLIP \cite{Cherti2023open_clip} ViT-L/14, pretrained on datacomp\_xl, as the VLM.
This model is fine-tuned using image-caption pairs, as described in \cref{sec:clip_finetuning}, using RADIal, CRUW, and nuScenes datasets.
We compare the original and fine-tuned VLM by evaluating retrieval performance for classes on the CRUW dataset, which are particularly relevant for autonomous driving. 
To compute the model predictions, the cosine similarity of the text and the image embeddings is computed. We rank the retrieved data samples by the cosine similarity values. The top-10 and top-100 retrieval metrics are listed in \cref{tab:rslm_top_retrieval}. 
The results show, that the fine-tuned VLM outperforms the original VLM on average. Performance is only worse on some classes like bicycle, likely due to those classes being underrepresented in the datasets used for fine-tuning.

\subsection{Radar spectra-language model}
\label{sec:experiment_radar_encoder}
%%------------ CLIP radar encoder -------------
\begin{table}[t]
\caption{Comparison of top-10 and top-100 precision scores for retrieval task for original VLM, fine-tuned VLM and \RSLM on CRUW dataset.}
    \tiny
\begin{center}
\begin{tabular}{l p{2cm} p{2cm} p{2cm} p{2cm}|p{2cm} p{2cm} p{2cm} p{2cm}}
                                   & \multicolumn{4}{c}{\textbf{Top 10}}                                                                                 & \multicolumn{4}{c}{\textbf{Top 100}}                                                                                \\[0.2cm] \midrule \\[2pt]
\multicolumn{1}{l|}{Label}         & \multicolumn{1}{c}{\makecell{Original \\ VLM}} & \multicolumn{1}{c}{\makecell{Fine- \\tuned \\ VLM}} & \multicolumn{1}{c}{\makecell{CNN \\ \RSLM}} & \multicolumn{1}{c|}{\makecell{FPN \\ \RSLM}} & \multicolumn{1}{c}{\makecell{Original \\ VLM}} & \multicolumn{1}{c}{\makecell{Fine- \\ tuned \\ VLM}} & \multicolumn{1}{c}{\makecell{CNN \\ \RSLM}} & \multicolumn{1}{c}{\makecell{FPN \\ \RSLM}} \\ \midrule

\multicolumn{1}{l|}{sidewalk}  & \multicolumn{1}{c}{1} & \multicolumn{1}{c}{1} & \multicolumn{1}{c}{1} &\multicolumn{1}{c|}{1} & \multicolumn{1}{c}{1} & \multicolumn{1}{c}{1} & \multicolumn{1}{c}{1} &\multicolumn{1}{c}{0.99}  \\
\multicolumn{1}{l|}{building}    & \multicolumn{1}{c}{1} & \multicolumn{1}{c}{1} & \multicolumn{1}{c}{1} & \multicolumn{1}{c|}{1} & \multicolumn{1}{c}{1} & \multicolumn{1}{c}{1} & \multicolumn{1}{c}{1} & \multicolumn{1}{c}{1}  \\
\multicolumn{1}{l|}{wall}   & \multicolumn{1}{c}{0.9} & \multicolumn{1}{c}{1} &  \multicolumn{1}{c}{0.4} & \multicolumn{1}{c|}{0.9} & \multicolumn{1}{c}{0.8} &\multicolumn{1}{c}{1} & \multicolumn{1}{c}{0.72} &  \multicolumn{1}{c}{0.86} \\
\multicolumn{1}{l|}{fence}  & \multicolumn{1}{c}{1} & \multicolumn{1}{c}{1}  & \multicolumn{1}{c}{0.9} & \multicolumn{1}{c|}{1} & \multicolumn{1}{c}{0.99} & \multicolumn{1}{c}{1} & \multicolumn{1}{c}{0.91} &  \multicolumn{1}{c}{0.96} \\
\multicolumn{1}{l|}{traffic light} & \multicolumn{1}{c}{0.3} & \multicolumn{1}{c}{0.2} &  \multicolumn{1}{c}{0.1} &  \multicolumn{1}{c|}{0.1}  & \multicolumn{1}{c}{0.57} & \multicolumn{1}{c}{0.17}  & \multicolumn{1}{c}{0.05} &  \multicolumn{1}{c}{0.06}  \\
\multicolumn{1}{l|}{traffic sign}  & \multicolumn{1}{c}{1} & \multicolumn{1}{c}{1}  &  \multicolumn{1}{c}{1} &  \multicolumn{1}{c|}{0.8} & \multicolumn{1}{c}{1} & \multicolumn{1}{c}{1} & \multicolumn{1}{c}{0.88} &  \multicolumn{1}{c}{0.87}\\
\multicolumn{1}{l|}{person}  & \multicolumn{1}{c}{0.2} & \multicolumn{1}{c}{1} &  \multicolumn{1}{c}{0} &  \multicolumn{1}{c|}{1} & \multicolumn{1}{c}{0.68} & \multicolumn{1}{c}{1} & \multicolumn{1}{c}{0.08} &  \multicolumn{1}{c}{0.95} \\
\multicolumn{1}{l|}{rider} & \multicolumn{1}{c}{0.8} & \multicolumn{1}{c}{1} &  \multicolumn{1}{c}{0.3} &  \multicolumn{1}{c|}{0.6} & \multicolumn{1}{c}{0.94} & \multicolumn{1}{c}{1} & \multicolumn{1}{c}{0.13} &  \multicolumn{1}{c}{0.4} \\
\multicolumn{1}{l|}{car} & \multicolumn{1}{c}{0.7} & \multicolumn{1}{c}{1} &  \multicolumn{1}{c}{1} &  \multicolumn{1}{c|}{1} & \multicolumn{1}{c}{0.84} & \multicolumn{1}{c}{0.92} & \multicolumn{1}{c}{1} &  \multicolumn{1}{c}{0.93} \\
\multicolumn{1}{l|}{truck} & \multicolumn{1}{c}{1} & \multicolumn{1}{c}{1} &  \multicolumn{1}{c}{0.6} &  \multicolumn{1}{c|}{0.7} & \multicolumn{1}{c}{1} & \multicolumn{1}{c}{0.85} & \multicolumn{1}{c}{0.28} &  \multicolumn{1}{c}{0.25} \\
\multicolumn{1}{l|}{bicycle} & \multicolumn{1}{c}{1} & \multicolumn{1}{c}{1} &  \multicolumn{1}{c}{0.7} &  \multicolumn{1}{c|}{0.4} & \multicolumn{1}{c}{1} & \multicolumn{1}{c}{0.26} &  \multicolumn{1}{c}{0.26}  & \multicolumn{1}{c}{0.2}     \\
\midrule
\multicolumn{1}{l|}{Mean} & \multicolumn{1}{c}{0.809}  & \multicolumn{1}{c}{0.927}  & \multicolumn{1}{c}{0.636}  & \multicolumn{1}{c}{0.773}  &  \multicolumn{1}{c}{0.893}  & \multicolumn{1}{c}{0.904}  & \multicolumn{1}{c}{0.574}  & \multicolumn{1}{c}{0.679} \\ 
\end{tabular}
\end{center}
\vspace{-0.5cm}
\label{tab:rslm_top_retrieval}
\end{table}
\textbf{Setup} 
We train a radar spectra encoder by matching the embedding of the corresponding image produced by the image encoder of the frozen VLM. Here we use the Open CLIP model fine-tuned to automotive scenes.
Separate encoders are trained for the range-Doppler spectra from RADIal dataset and range-azimuth spectra from CRUW dataset. We trained the CNN and FPN radar encoder with mean squared error (MSE) loss for matching the embeddings.
%%------------ Evaluation -------------

\textbf{Evaluation of \RSLM}
The trained {\RSLM}s are evaluated on a retrieval task as described above. The CLIP text encoder and our trained radar encoder are used to compute the radar spectra-language model predictions. The retrieved data samples are ranked by the cosine similarity values.

\begin{figure}
    \begin{subfigure}[t]{0.49\linewidth}
        \includegraphics[width=\linewidth]{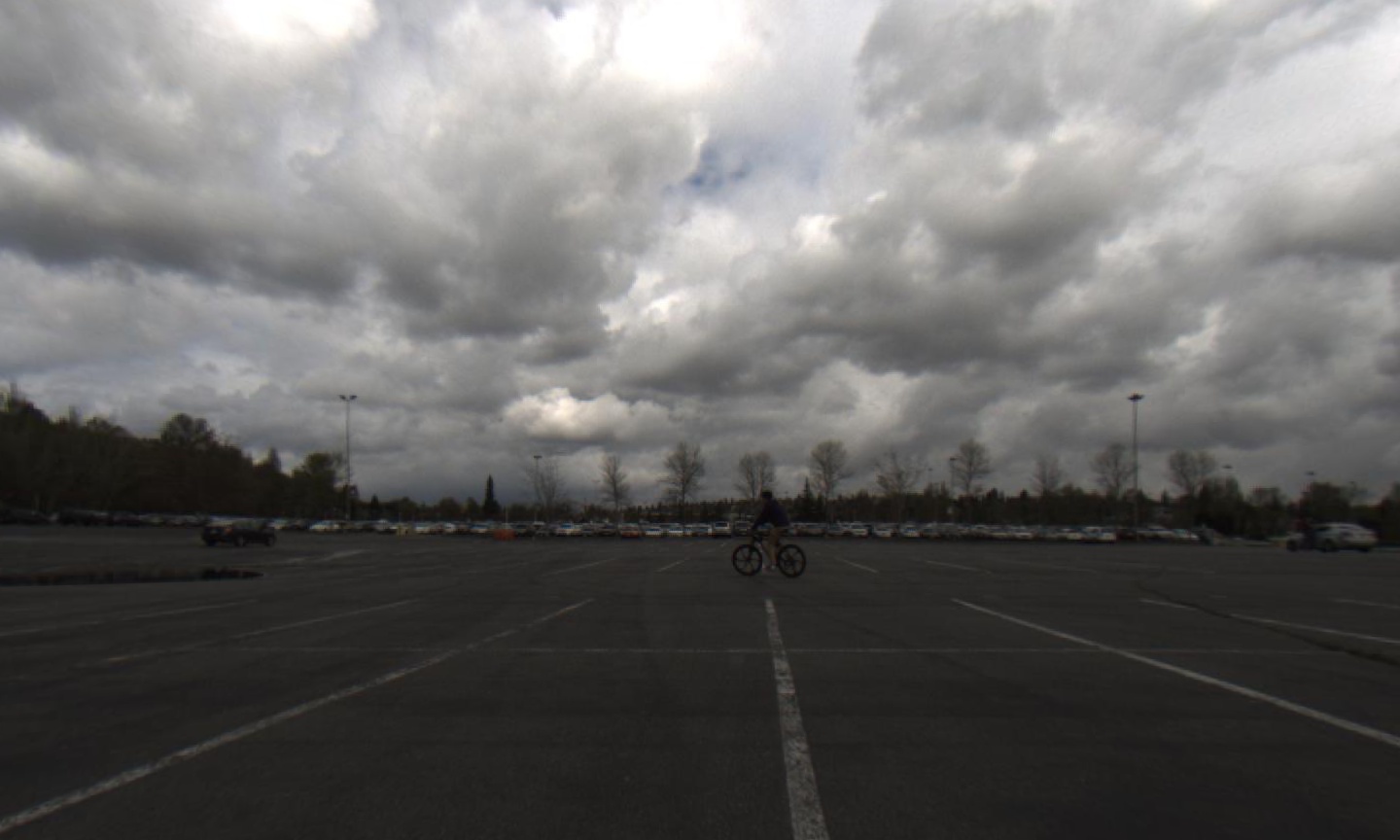} %
        \caption{Parking lot with many cars}
        \label{fig:retrieval_parking}
    \end{subfigure}%
    \hfill
    \begin{subfigure}[t]{0.49\linewidth}
        \includegraphics[width=\linewidth]{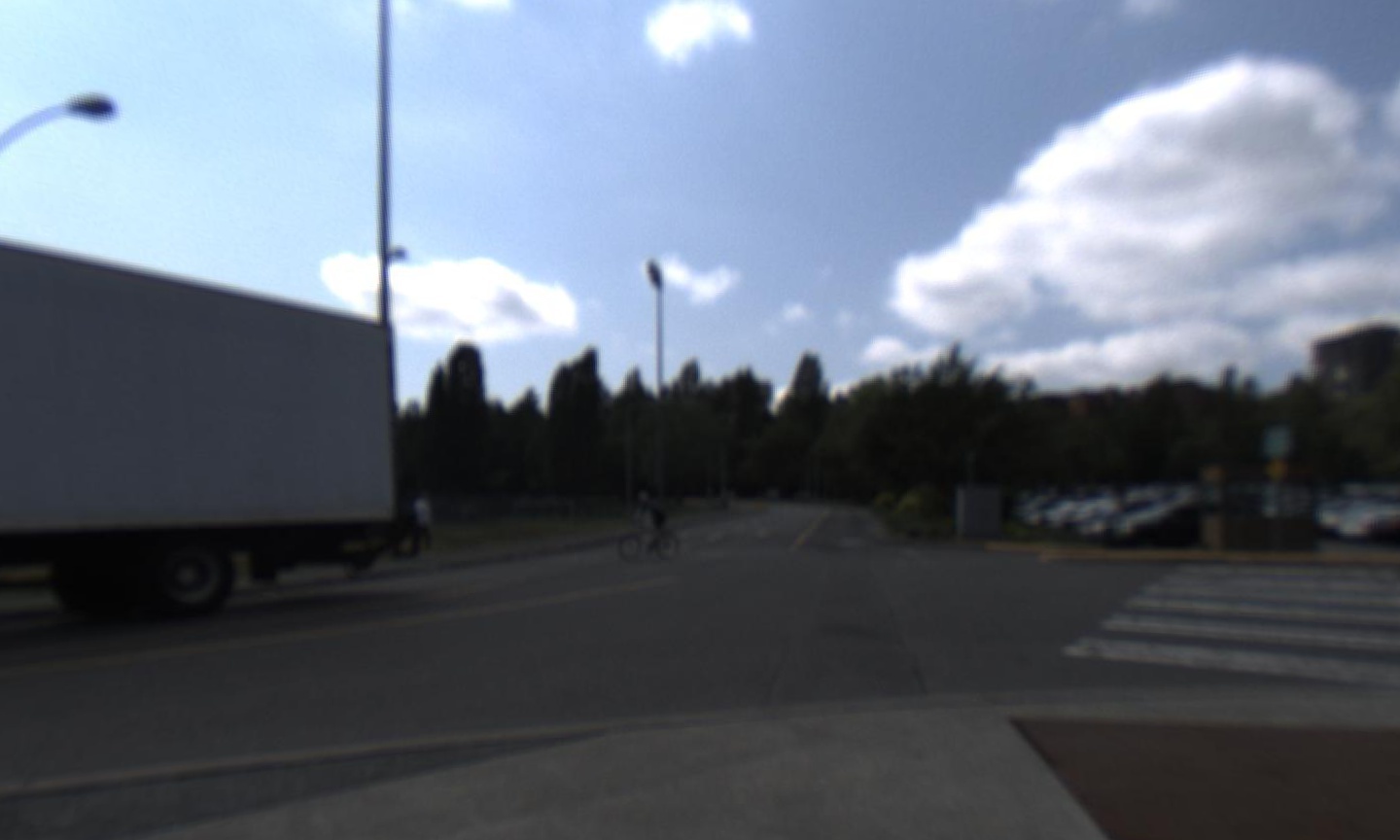}
        \caption{truck cruising confidently on the open road}
        \label{fig:retrieval_truck}
    \end{subfigure}
    \caption{Data retrieval using the trained RSLM. The corresponding images are shown for visualization only, they are not used for data retrieval. The used query appears in the caption of each image.}
    \label{fig:retrieval_captions}
    \vspace{-11pt}
\end{figure}

\textbf{Results} 
Retrieval performance is shown in \cref{tab:rslm_top_retrieval}. The FPN radar encoder outperforms the CNN radar encoder for most of the classes and by mean top-10 and top-100 accuracy. For person prompt, FPN achieves better results than the original VLM, due to fine-tuning. Thus, the \RSLM can be successfully applied to retrieval tasks.
In \cref{fig:retrieval_captions} images are shown, which correspond to the retrieved spectra with maximal cosine similarity value for the given caption. It shows, that the \RSLM can retrieve objects and scenes like parking lots and trucks, which were not presented in the ground truth.
This shows, that radar spectra and language can be succesfully connected using the \RSLM.

\subsection{Object detection and free-space segmentation with \RSLM radar embeddings}
\label{sec:experiment_object_detection} 
%%------------ Detection with CLIP radar encoder -------------
To evaluate the benefit of the learned radar embeddings of the RSLM, we compare a) the baseline network ``FFT-RadNet" \cite{Rebut2021radial} with b) the baseline network including the radar encoder of the RSLM ``FFT-RadNet + RSLM encoder'', \cf \cref{sec:det_clip}.
We were not able to reproduce the results of ``FFT-RadNet" reported in \cite{Rebut2021radial} using the provided hyperparameters. Therefore, we provide the results of our training to be able to compare it to the model including the RSLM radar encoder. 

\textbf{Metrics}
We evaluate the models on the RADIal dataset and use the same metrics as in \cite{Rebut2021radial}: For evaluating the object detection task the mean average precision (mAP), mean average recall (mAR), and F1-score are computed. For free space segmentation the intersection over union (IoU) is used.

\begin{figure}
    \centering
    \begin{subfigure}[c]{0.45\linewidth}
        \centering    
        \includegraphics[width=\linewidth]{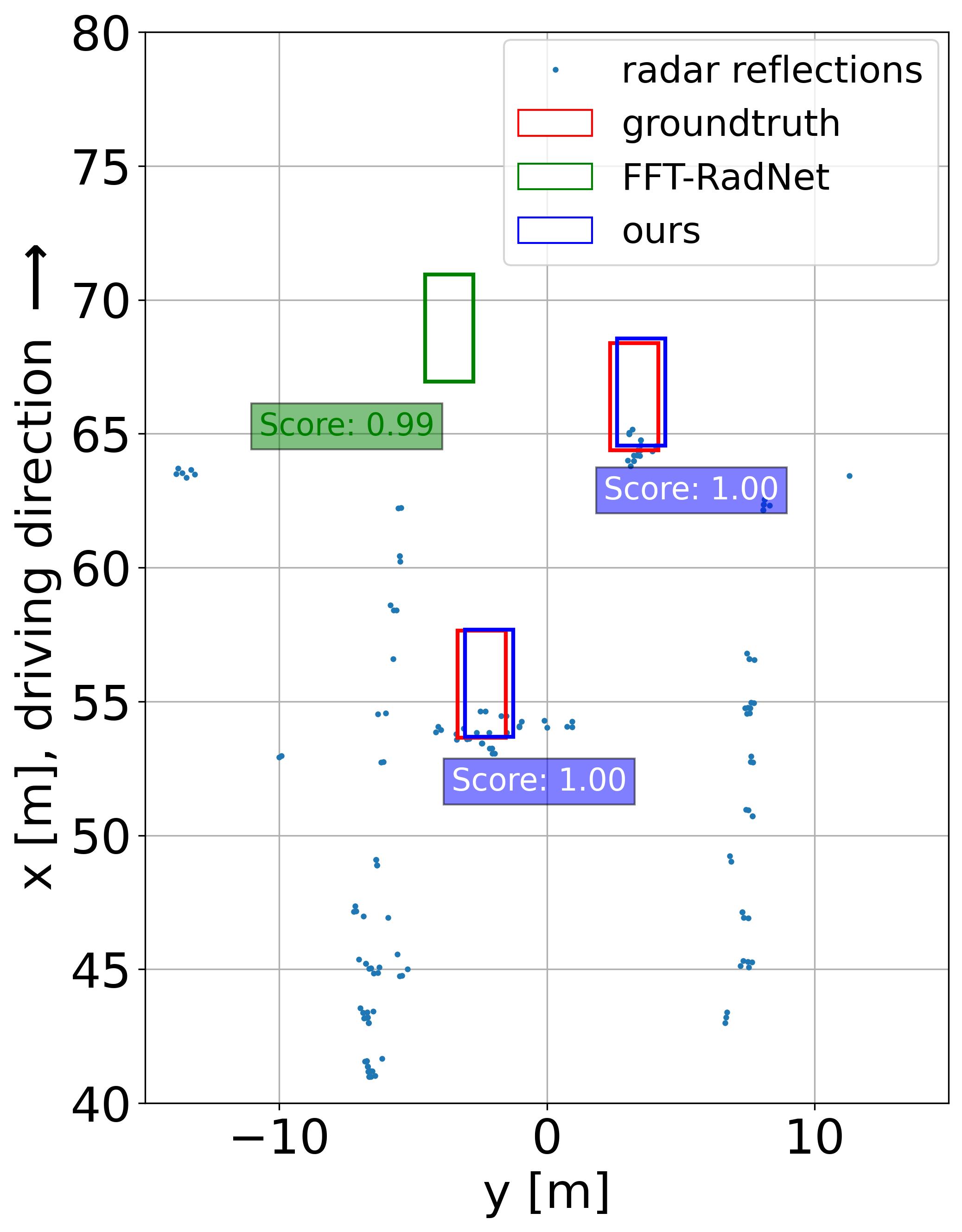}
    \end{subfigure}
    \hfill
    \begin{subfigure}[c]{0.53\linewidth}
        \centering    
        \includegraphics[width=\linewidth, clip, trim={8cm 2cm 8cm 2cm}]{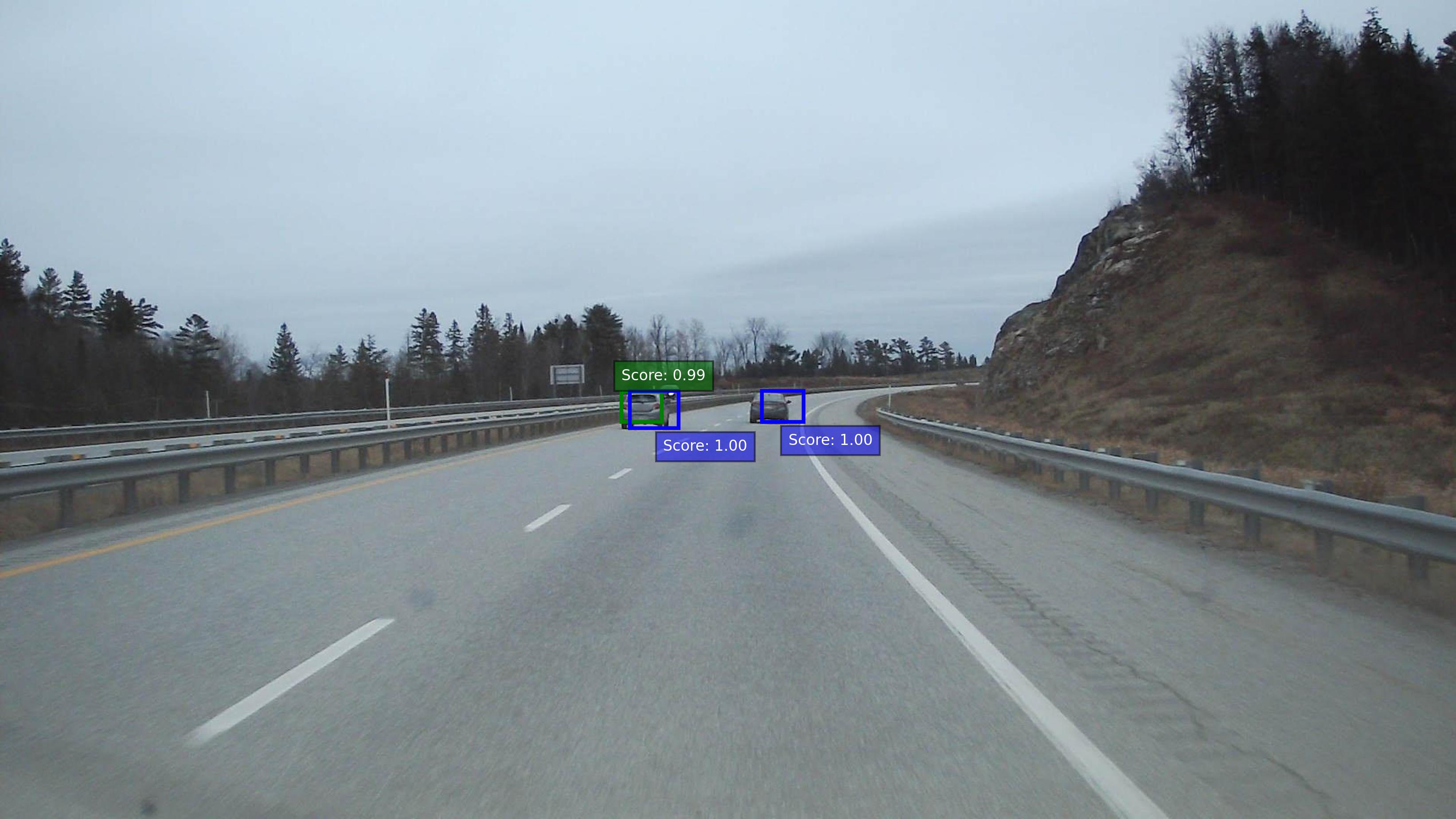}
        %trim: left lower right upper
    \end{subfigure}   
   \caption{Detection results of FFT-RadNet (green) and our proposed network ``FFT-RadNet + RLSM encoder'' (blue). The bounding box prediction of FFT-RadNet is displaced \wrt the ground truth (red), whereas the predictions of our model align well with the ground truth. Confidence score equals 1.0. Left: Bounding boxes in Cartesian coordinates, radar point clouds displayed for reference only. Note that the models work on spectral data. Thus objects might even be predicted at locations where no radar point clouds are visible. Right: Bounding boxes projected on image.}
   \label{fig:det_example}
   \vspace{-8pt}
\end{figure}

\begin{figure}
    \centering
    \begin{subfigure}{0.5\linewidth}
        \centering    
        \includegraphics[width=\linewidth, clip, trim={5cm 0 5cm 6cm}]{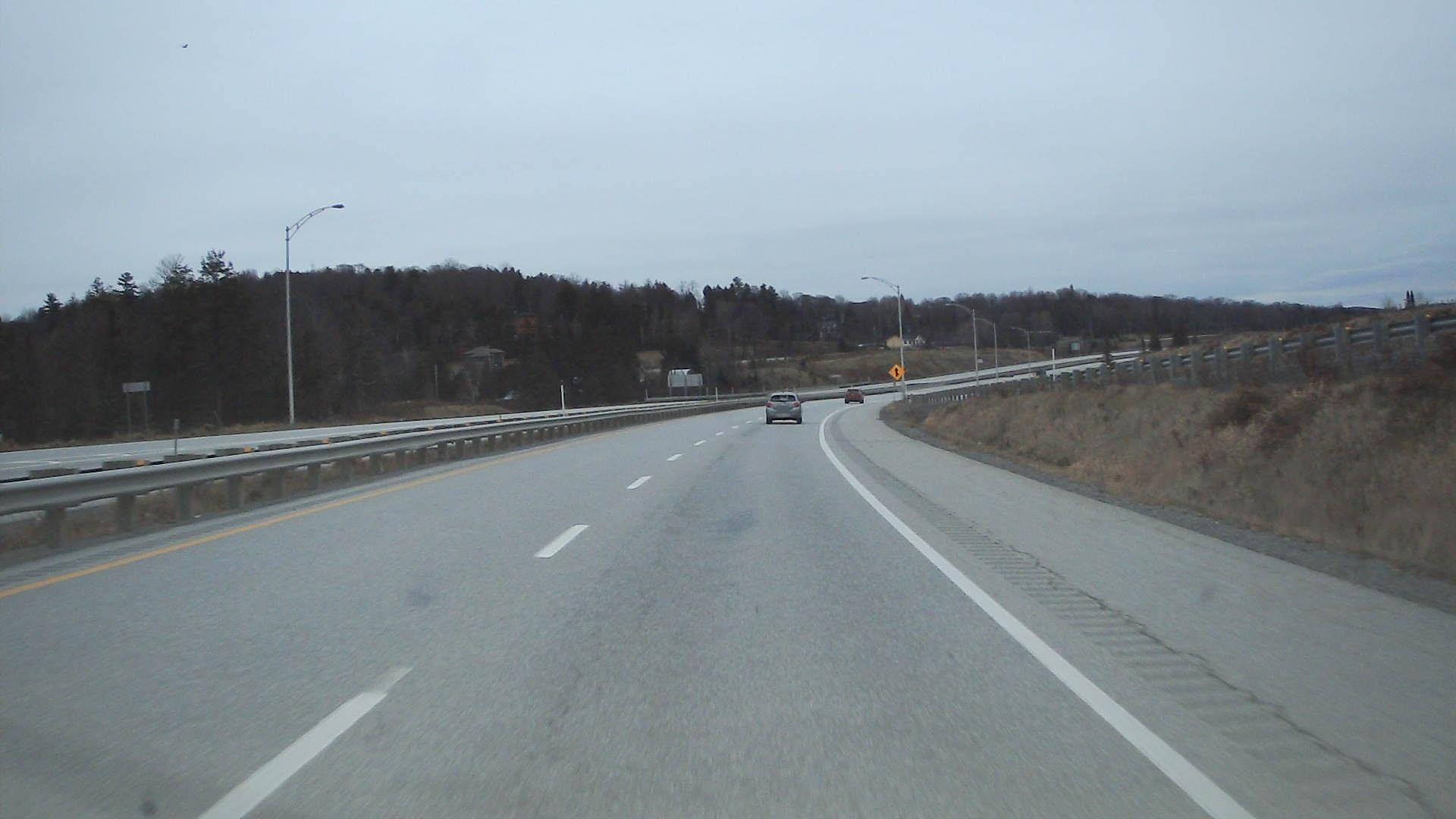}
        %trim: left lower right upper
        \caption{RGB}
    \end{subfigure}
    % \hfill
    \par\medskip
    \begin{subfigure}{0.45\linewidth}
        \centering    
        \includegraphics[width=\linewidth]{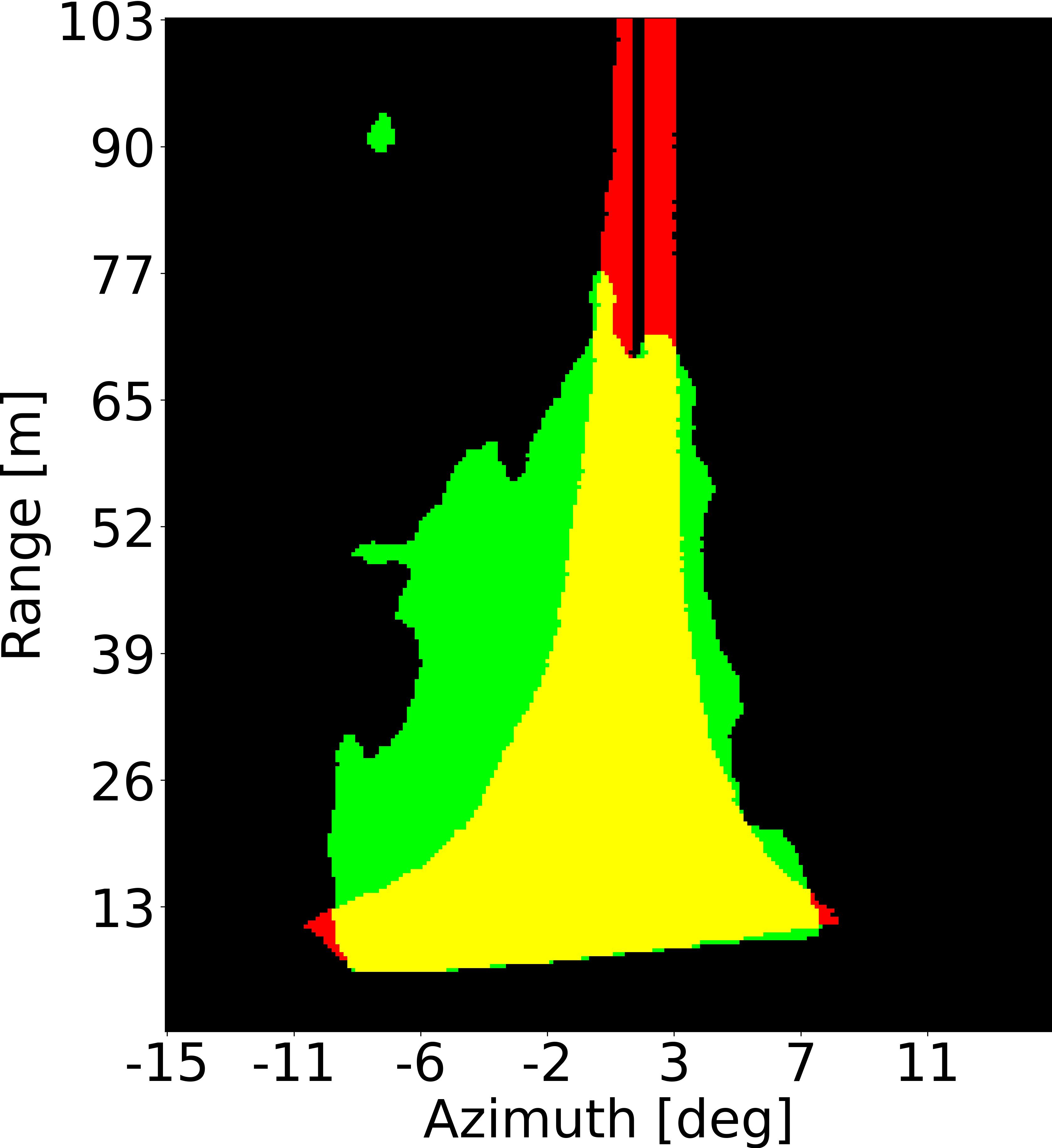}      
        \caption{FFT-RadNet}    
    \end{subfigure}
    % \hfill
    \begin{subfigure}{0.45\linewidth}
        \centering    
        \includegraphics[width=\linewidth]{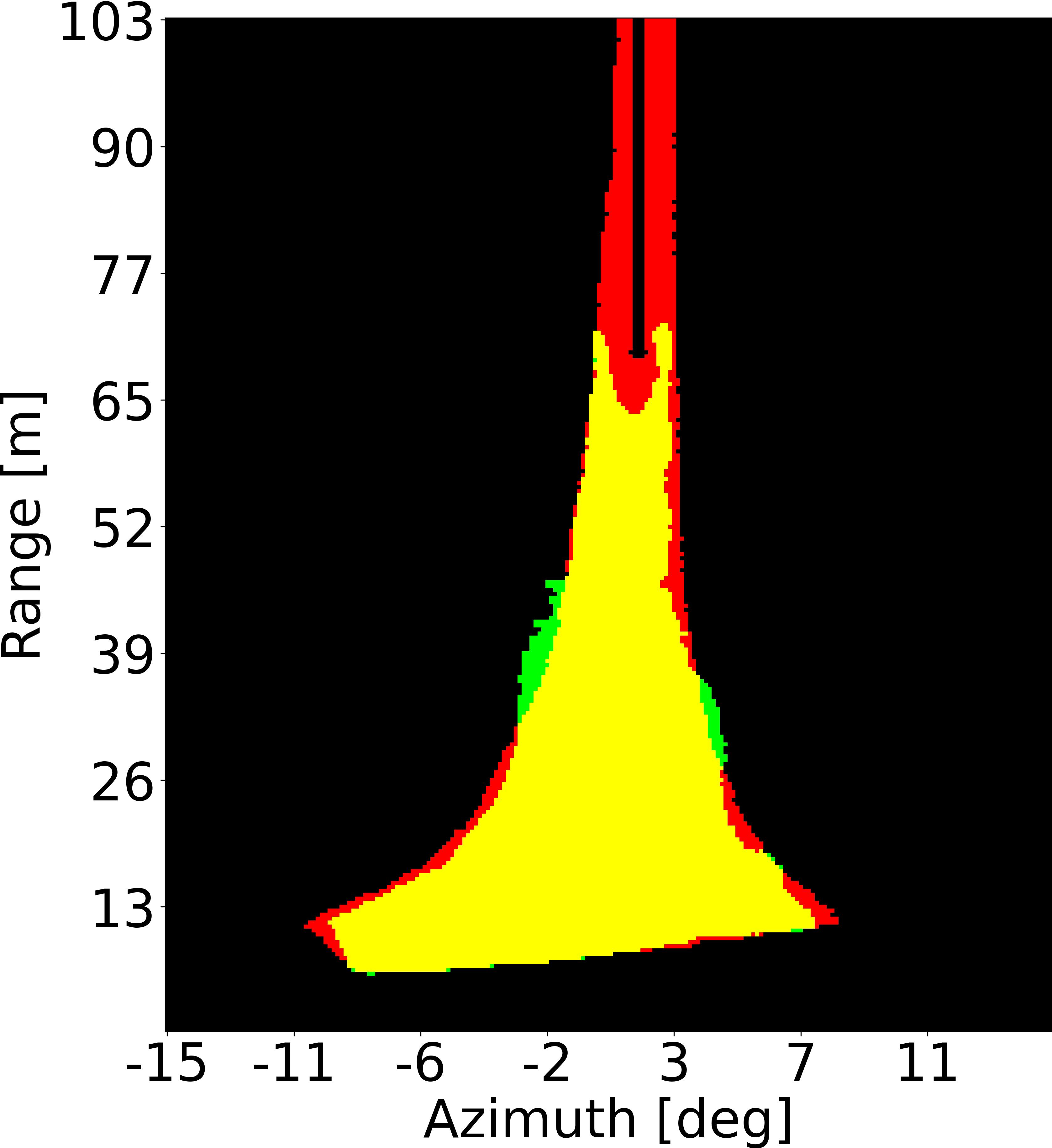}      
        \caption{Ours} 
    \end{subfigure}
   \caption{Example of segmentation result of FFT-RadNet and our network ``FFT-RadNet + RLSM encoder''. Red color denotes ground truth open driving space, green color represents free-space predicted by the corresponding model,
and yellow color denotes the intersection of ground truth and predicted drive-able space. 
 The predictions of our proposed method ``FFT-RadNet + RLSM encoder'' are better aligned to the ground truth. Note that the models use radar spectra only as input.}
   \label{fig:seg_example}
   \vspace{-2pt}
\end{figure}

\textbf{Results}
\cref{tab:detection} summarizes the results. Using the pre-trained RLSM radar encoder (``FFT-RadNet +
RLSM encoder'') improves object detection (mAP and F1-score) and segmentation performance (IoU). mAR is very similar for both models. Thus, simply injecting the pre-trained radar spectra embeddings of the RSLM encoder improves the performance. Note that no additional labeled data is necessary to pre-train the RLSM encoder. 
Visualizations for detection and segmentation results can be found in \cref{fig:det_example} and \cref{fig:seg_example}.

\begin{table}[t]
     \small
     \caption{Comparison of the baseline model ``FFT-RadNet'' with our proposed model ``FFT-RadNet + RSLM encoder''. Simply injecting the pre-trained radar spectra embeddings of the RSLM encoder improves object detection (mAP, mAR, F1) and segmentation (IoU) performance. Not that pre-training the RSLM encoder does not require any additional ground truth data.}
     
         \begin{tabular*}{\linewidth}{@{\extracolsep\fill}l | c c c c}
             \toprule%
             Model  & mAP (\%) & mAR (\%) & F1 (\%) & IoU (\%) \\ 
             \midrule
             FFT-RadNet$^{(*)}$  & 88.8 $\pm$ 1.7 & 81.2 $\pm$ 1.8 & 84.2 & 67.3 $\pm$ 1 \\ 
             \makecell{FFT-RadNet + \\RLSM encoder} & \textbf{90.7} $\pm$ 1.1 & 81.8 $\pm$ 2 & \textbf{86.0} & \textbf{71.2} $\pm$ 2.3  \\
             \bottomrule
         \end{tabular*}
         \text{$\mathbf{(*)}$ FFT-RadNet architecture from \cite{Rebut2021radial}, trained by us.}
     \label{tab:detection}
     \vspace{-0.2cm}
\end{table}

\begin{table*}[t]
    \caption{Ablation studies for detection (mAP, mAR, and $F_1$-score) and segmentation task (IoU). The different model architectures are described in \cref{sec:experiment_object_detection}. The ``frozen enc'' and ``from scratch'' models, achieve the best results. Note that training the ``frozen enc'' model doesn't require any additional ground truth data.}
        \begin{tabular*}{\textwidth}{@{\extracolsep\fill}l | c c c c  | c c c c}
        \toprule%
            Model  & \makecell{Detect \\ backbone }  & \makecell{Radar \\ enc}  & \makecell{\RSLM \\weights} & \makecell{Fine-tuned \\ enc} & mAP (\%) & mAR (\%) & F1 (\%) & IoU (\%) \\ 
            \midrule
            frozen enc  &  + & + & + & - & \textbf{90.7} $\pm$ 1.1 & 81.8 $\pm$ 2 & \textbf{86.0} & 71.2 $\pm$ 2.3  \\
            fine-tuned enc & + & + & + & + & 90.4 $\pm$ 1.2 & 81.4 $\pm$ 2.1 & 85.6 & 69.9 $\pm$ 2.6\\
            only-frozen enc & - & + & + & - & 0.1 $\pm$ 0 & 2.4 $\pm$ 0.6 & 0.1 & 55 $\pm$ 16.7\\
            only fine-tuned enc  & - & + & + & + & $ 0.0\pm$ 0& 2.7 $\pm$ 1.1 & 0 & 59.1 $\pm$ 9.9\\
            from-scratch & + & + & - & + & 88.1 $\pm$ 2.8 & \textbf{82.9} $\pm$ 0.7 & 85.4 & \textbf{72.6} $\pm$ 1.9\\
        \bottomrule
        \end{tabular*}
        \text{$\mathbf{(*)}$ FFT-RadNet architecture from \cite{Rebut2021radial}, trained by us.}
    \label{tab:ablation}
    \vspace{-0.1cm}
\end{table*}

\textbf{Ablation Study}
For better understanding of the model and training methods, we conduct an ablation study and compare the following models: ``baseline'' network is FFT-RadNet \cite{Rebut2021radial}, ``frozen-enc'' denotes the network ``FFT-RadNet + RSLM encoder'' with the frozen, pre-trained radar encoder. ``fine-tuned enc'' is the same network as ``frozen-enc'', however, the radar encoder is fine-tuned on the last 10 epochs. ``only frozen enc'' includes the pre-trained radar encoder, radar adapter, detection and segmentation head only. ``only fine-tuned enc'' is the same as ``only frozen enc'', but the radar encoder is fine-tuned on last 10 epochs. ``from-scratch'' is a random initialized network with the "frozen-enc" architecture.

\cref{tab:ablation} summarizes object detection and segmentation performance for the models described above. 
The columns of the table correspond to model features: ``detect backbone'' signifies the use of the detection backbone (MIMO pre-encoder and FPN Radar backbone) of FFT-RadNet. ``radar enc'' denotes the incorporation of the radar encoder from \RSLM. ``\RSLM weights'' indicates the usage of weights from the \RSLM model for the radar encoder; otherwise, it is randomly initialized. ``Fine-tuned enc'' signifies that the radar encoder was fine-tuned during detection training; otherwise it is frozen. Results in the table exhibit performance improvements when adding frozen radar embeddings into the model architecture.
This enhancement is observed in both detection and free-space segmentation tasks, as compared to the baseline model without embeddings (``frozen enc'' and ``fine-tuned enc'' vs. ``baseline'').
Fine-tuning the radar encoder does not improve object detection or segmentation performance.
Furthermore, the model ``from-scratch'' with the same architecture as the ``frozen-enc'' variant, exhibits slightly higher IoU scores for free-space segmentation and similar detection performance compared to the ``frozen-enc'' model.
In contrast, models that exclusively incorporate the radar encoder component of \RSLM (``only-frozen enc'', ``only fine-tuned enc''), whether frozen or trained during the last 10 epochs, do not successfully accomplish the detection task.

This shows, that using the pre-trained radar encoder from \RSLM in addition to the detection backbone improves performance in downstream tasks (``frozen enc'' vs. ``baseline''), \ie learning the feature embeddings is helpful.
Note, that the pre-training does not use any labeled radar spectra data, and the weights of the radar encoder lead to similar performance as weights trained in a fully supervised manner (``frozen enc'' vs. ``from-scratch'').  We emphasize that improvements are achieved with the same hyperparameters as the baseline model by just adding the \RSLM radar encoder.

\textbf{Discussion and Future Directions}
The proposed \RSLM relies on a pre-trained vision-language model, and therefore depends on the quality of the caption-image pairs the underlying model was trained on. More captions would help to fine-tune the corresponding VLM, yielding a better \RSLM. This performance dependence is an obvious limitation of the \RSLM, as the VLM has a limited performance on some fine-grained cases, e.g. it often cannot recognize traffic signs.

Our results demonstrate that the proposed \RSLM can learn relevant features for scene retrieval. While only scene-level descriptions have been considered in this paper, object-, or region- level descriptions (a 3D variant of e.g. \cite{Li2021glip,Zhong21cvpr-RegionCLIP,Yao23cvpr-DetCLIPv2}) would also be beneficial in future work. 

Our experiments show that the learned features are relevant for downstream-tasks. We emphasize that the observed performance boost is obtained without the need for any additional labeled data, only by making use of image-radar pairs.

Radar measurements are not significantly affected by bad weather conditions or time of day, which is one main advantage.
Therefore, while the proposed \RSLM was trained on images that are taken at daytime, it can be expected to work as well in ``rainy'', and ``night'' scenarios. However, this is not possible to verify since images in difficult conditions are not available for the considered datasets. New publicly released datasets with difficult weather conditions, such as the recently available \cite{paek-2022-k-radar-dataset}, will be beneficial.
As another future application, the proposed model might be used for radar data generation.

\section{Conclusion}
\label{sec:conclusion}
We developed a radar spectra-language model (\RSLM), to the best of our knowledge the first such model, for automotive scenes.
Our method makes use of vision-language models (VLMs), which we first fine-tuned on automotive image data to improve their performance.
We investigated the semantic content of radar spectra, by querying the \RSLM with text descriptions and evaluating radar scene retrieval.
In this way the model can even be used to query for different object types, for which no corresponding labels exist in the dataset.
Moreover, the proposed methods overcomes the scarcity of labeled radar spectra data, since no labeled radar data is needed to train the \RSLM.
Finally, we showed that the performance in downstream tasks can be improved by injecting radar feature embeddings from the \RSLM into a detection and segmentation model.

\bibliographystyle{IEEEtran}
\bibliography{IEEEabrv,refs}

\end{document}